\documentclass[sigconf]{acmart}

\acmYear{2026}\copyrightyear{2026}
\setcopyright{cc}
\setcctype[4.0]{by}
\acmConference[CAIS '26]{Proceedings of the 1st ACM Conference on Agentic and AI Systems}{May 26, 2026}{San Jose, CA, USA}
\acmBooktitle{Proceedings of the 1st ACM Conference on Agentic and AI Systems (CAIS '26), May 26, 2026, San Jose, CA, USA}
\acmDOI{10.1145/3786335.3813228}
\acmISBN{979-8-4007-2415-2/26/05}

\AtBeginDocument{%
  }

\begin{document}

\title{A Compound AI Agent for Conversational Grant Discovery}

\author{Zhisheng Tang}
\affiliation{%
  \institution{University of Southern California \& GRAIL}
  \country{United States}}
\email{zhisheng@isi.edu}

\author{Mayank Kejriwal}
\affiliation{%
  \institution{University of Southern California \& GRAIL}
  \country{United States}}
\email{kejriwal@isi.edu}

\begin{abstract}
Research funding discovery remains fundamentally fragmented: researchers navigate disparate agency portals (e.g., in the United States, NSF, NIH, DARPA, Grants.gov, and many others) with heterogeneous interfaces, search capabilities, and data schemas. We present a compound AI system that unifies this landscape through two tightly coupled components: (1) an aggregation layer that autonomously collects, normalizes, and indexes almost 12,000 federal and nonprofit opportunities from fragmented sources via LLM-equipped browser agents, maintaining a biweekly-updated unified database; and (2) an agentic ReAct-based query processing layer that interprets research context (including from PDF documents) and employs hybrid search combining a structured index with selective web search to retrieve relevant opportunities - while avoiding LLM hallucination. The conversational interface supports iterative refinement through multi-turn interactions, allowing researchers to progressively apply constraints without reformulating their core research description. Results stream in real time with full transparency of intermediate reasoning, enabling appropriate calibration of user trust. Currently used by almost 3,000+ users, our approach demonstrates the feasibility of compound AI in reducing grant discovery time from 30--45 minutes (manual, fragmented portal searches) to under 10 minutes (unified, conversational search).
\end{abstract}

\maketitle

\section{Background and Related Work}

Securing research funding is fundamental to academic innovation~\cite{sohn2020secrets}, yet the grant proposal process remains fragmented and time-consuming. Researchers must identify opportunities across dozens of separate agencies and foundations while simultaneously navigating demanding proposal writing. The funding discovery phase itself presents a critical bottleneck: in the United States, federal agencies like the National Science Foundation (NSF) and the Department of Energy (DOE), as well as nonprofit foundations operate independent portals with distinct search interfaces. Existing grant discovery relies on some combination of information and instructions in these fragmented portals (\url{https://www.grants.gov}, NSF FastLane\footnote{\url{https://fastlane.nsf.gov}}, NIH REPORTER\footnote{\url{https://reporter.nih.gov}}) that require researchers to manage multiple interfaces, many of which have little AI support for efficient insight-gathering.

Recent advances in large language models (LLMs) have demonstrated significant promise for supporting scientific writing and research workflows~\cite{huang2023role, song2023enhancing, lavric2023brainstorming}. LLM-based tools facilitate drafting, editing, and stylistic refinement, particularly benefiting researchers who face barriers in scientific communication~\cite{katsnelson2022poor, amano2023manifold}. Building on these successes, specialized frameworks for grant writing have emerged~\cite{seckel2024ten, godwin2024grant}, recognizing that effective AI assistance benefits from domain-specific support. However, a naive approach using a general-purpose LLM (e.g., ChatGPT) with web search reveals fundamental limitations: web search cannot reliably access fragmented agency databases, and LLMs hallucinate details about programs they have not been trained on, introducing factual errors~\cite{alkaissi2023artificial, mcgowan2023chatgpt} and surfacing outdated information. Without a unified, normalized index into which search queries on grants can be grounded, such approaches cannot guarantee consistency across heterogeneous sources.

Recent work demonstrates that effective AI assistance for specialized domains like grant writing benefits from system design that integrates LLMs with structured data and domain-specific support~\cite{seckel2024ten, godwin2024grant}. Research on AI-assisted systems confirms that LLMs are most effective as supplements to human judgment, and that appropriate system design is critical for enabling effective human-AI collaboration~\cite{amershi2019guidelines, liao2022designing}. Interface design that exposes system reasoning and intermediate results can calibrate user trust and reduce over-reliance on automated suggestions~\cite{kizilcec2016much, bucinca2021trust}.

% A \textit{compound AI system} 

\begin{figure}
\centering
\includegraphics[width=0.8\columnwidth]{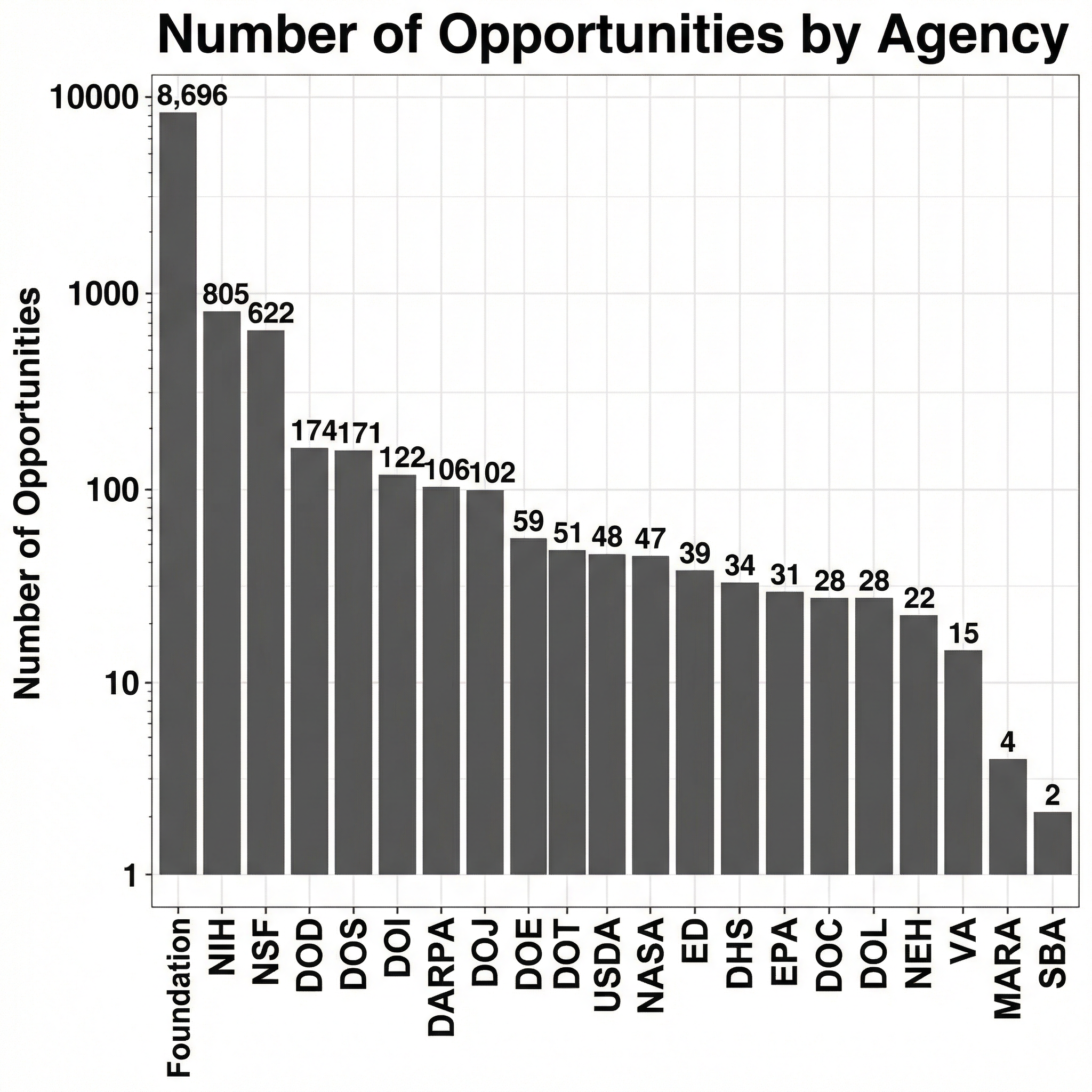}
\caption{Distribution of funding opportunities across U.S. federal agencies and nonprofit foundations in our unified index. The Foundation category (8,696 opportunities) and NSF (805) dominate, followed by traditional research agencies. This diversity underscores the fragmentation challenge that motivates our unified discovery system.}
\label{fig:opportunities}
\end{figure}

We demonstrate a compound AI system that addresses these limitations through two tightly coupled components. By \emph{compound}, we mean a design that integrates multiple components working in sequence or parallel to solve a complex task, rather than relying on a single monolithic model~\cite{zaharia2024compound}. Such systems combine structured data pipelines, retrieval modules, and language models to achieve higher accuracy and reliability while avoiding common pitfalls like hallucination and staleness. The first component in our system is an aggregation pipeline that autonomously scrapes Grants.gov, DARPA, and foundation websites on a biweekly schedule, extracting and normalizing grant information into a unified, structured index of 11.8K opportunities (see \autoref{fig:opportunities}). The second component is an agentic LLM that employs a ReAct-based framework with two tools: \texttt{search\_index} and \texttt{web\_search}. This enables intelligent discovery of funding opportunities. Critically, users can upload PDF documents (grant proposals, papers, research summaries), and the system passes the full text to the language model as context, enabling automatic extraction of domain-specific keywords without manual specification. The agent then searches the fresh, unified index and complements with web search when needed, streaming results incrementally to the user.

By separating the indexing layer (which maintains freshness and accuracy through biweekly updates) from the query processing layer (which provides natural language interaction and PDF context understanding), the system avoids the fundamental trade-offs that constrain naive LLM applications. This architecture directly addresses the tension in AI-assisted research: leveraging language understanding while avoiding hallucination and staleness. The system has been deployed as a web-based platform and currently serves 3,000+ users, validating the practical value of compound AI for funding discovery.

\section{System Architecture}

\begin{figure}
\centering
\includegraphics[width=\columnwidth]{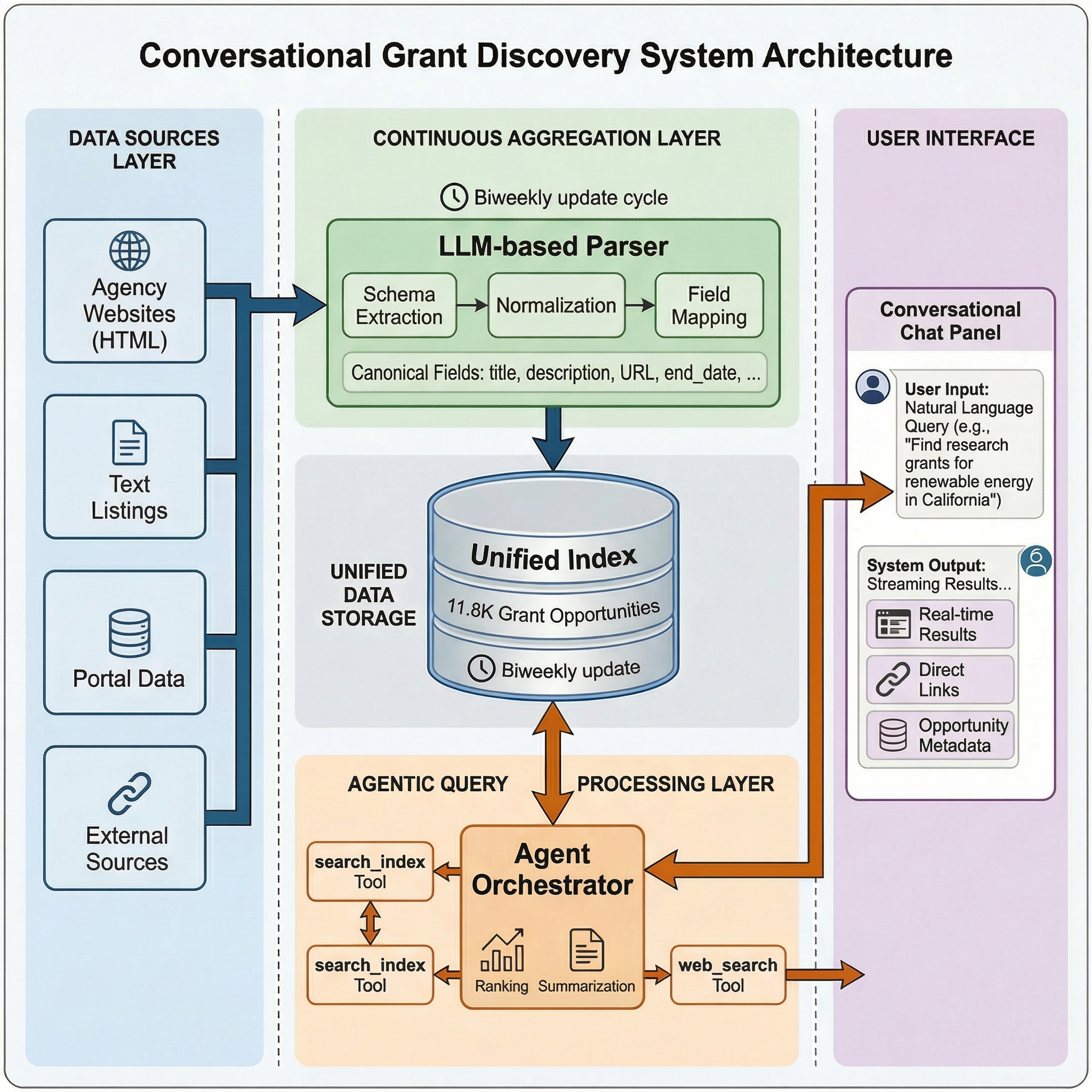}
\caption{System architecture of the \textit{Conversational Grant Discovery} system. Data flows from heterogeneous sources through the aggregation layer (LLM-based parsing and normalization) into a unified index. Users interact with the conversational interface, which feeds queries through the agentic query processing layer equipped with search\_index and web\_search tools, returning ranked, streamed results with direct agency links.}
\label{fig:architecture}
\end{figure}

We implement a compound AI system~\cite{zaharia2024compound} composed of two tightly coupled layers: an aggregation pipeline that maintains a fresh, unified index across fragmented funding sources, and an agentic query processor that supports exploratory discovery through natural language conversation. Figure~\ref{fig:architecture} illustrates the end-to-end system architecture, showing how data flows from heterogeneous sources through aggregation into a unified index, which the agentic layer queries to return ranked results to users.

\paragraph{Aggregation Layer}

The aggregation layer ingests data from multiple sources including federal agencies e.g., Grants.gov, DARPA, NIH, NSF, and nonprofit and foundation websites. These sources employ heterogeneous data formats and structures, requiring distinct data collection strategies. For federal agencies, we gather all opportunity pages and deploy an \textit{LLM-equipped browser agent} to autonomously discover and extract information. For nonprofit and foundation websites, we first retrieve the main webpage URL from tax records, then use Firecrawl to enumerate all discoverable URLs under that domain. A small language model ranks these URLs to identify the top ten most likely to contain grant opportunity information. Each of these ten URLs is then processed using the same LLM-browser methodology applied to federal sources.

LLM-based parsing converts unstructured and semi-structured source data into normalized records. Rather than encoding brittle extraction rules for each source variant, we pass the raw HTML or text through an LLM with a schema specification, requesting extraction of several canonical fields including \texttt{title}, \texttt{description}, \texttt{url}, and \texttt{end\_date}. This approach gracefully handles structural variation across agency websites and tolerates malformed or missing fields. Biweekly updates balance freshness with computational cost, ensuring researchers encounter opportunities well before typical submission deadlines.
Parsed records are synced to Algolia\footnote{https://www.algolia.com/}, a search engine that supports diverse search criteria including keyword-based and semantic queries. The unified schema eliminates the cognitive overhead of navigating multiple agency interfaces: all opportunity metadata is uniformly accessible; and search capabilities capture conceptually relevant opportunities without requiring researchers to know agency-specific terminology.

\paragraph{Agentic Query Processing Layer}

The query processing layer employs a ReAct-based agentic framework in which an LLM is equipped with two tools: (1) \texttt{search\_index}, which queries the unified vector database, and (2) \texttt{web\_search}, which retrieves current information from the public web. The agent's decision logic is explicit: its goal is to find relevant grant opportunities satisfying the user's query, and it decides which tools to invoke and in what order.

Users submit requests via natural language conversation or by uploading PDF documents - a workflow that we will also mirror in the live demonstration plan described in Section \ref{demo}. When a PDF is provided (e.g., a grant proposal draft or research paper), the full text is extracted and passed directly to the LLM as context, which enables the agent to understand the researcher's detailed scientific background, specific methodologies, and research goals without requiring manual summarization or keyword specification.

The agent follows a canonical strategy: it first invokes \texttt{search\_index} to retrieve results from the biweekly-updated, structured index of 11.8K opportunities, using keyword matching across titles, descriptions, deadlines, and URLs to identify opportunities matching user-specified or agent-extracted terms. This ensures results are factually grounded in explicitly verified, up-to-date grant information—avoiding the hallucinations and outdated information that plague naive LLM-based approaches. If initial index results are sparse or if the agent determines that very recent funding announcements may have been posted since the last biweekly update, it then invokes \texttt{web\_search} to find complementary results. This hybrid strategy combines the strengths of both approaches: the index provides comprehensive, normalized, structured data that web search cannot reliably access across fragmented agency portals, while web search captures very recent postings.

Results are streamed incrementally to the user, with the first token appearing within 2 seconds of the user's request. As results are identified, ranked, and summarized by the agent, they are immediately sent to the client, enabling users to begin reviewing opportunities while the agent continues searching. Each result presents opportunity metadata (title, agency, deadline, URL) grounded in direct hyperlinks to official agency pages, ensuring all suggestions are factual and verifiable.

\section{Interface and Interaction Patterns}

\paragraph{Conversational Grant Discovery}

The user interface is a global chat panel where researchers describe their work in natural language. Rather than navigating complex search forms or learning database-specific query syntax, users simply explain their research interests in conversational language. The agent interprets the natural language description, extracts domain-specific keywords, queries the unified index, and presents relevant opportunities with program titles, descriptions, deadlines, and direct links to funding agency pages.

This interface design reflects a fundamental insight about grant discovery: the process is inherently exploratory and iterative. Researchers often do not know in advance which funding sources or program types are relevant to their work. A traditional form-based search interface requires researchers to predict their search terms and constraints upfront, which is cognitively demanding when the landscape is unfamiliar. In contrast, the conversational interface enables progressive exploration, allowing constraints to be applied incrementally (deadline, collaboration requirements, funding source, research scope) without requiring researchers to reformulate their core research description with each refinement.

\subsubsection*{Concrete Use Case}

\begin{figure*}
\centering
\begin{minipage}[t]{0.48\textwidth}
\centering
\textbf{Pane A: Grant Discovery System}\\[0.5em]
\includegraphics[width=0.95\textwidth]{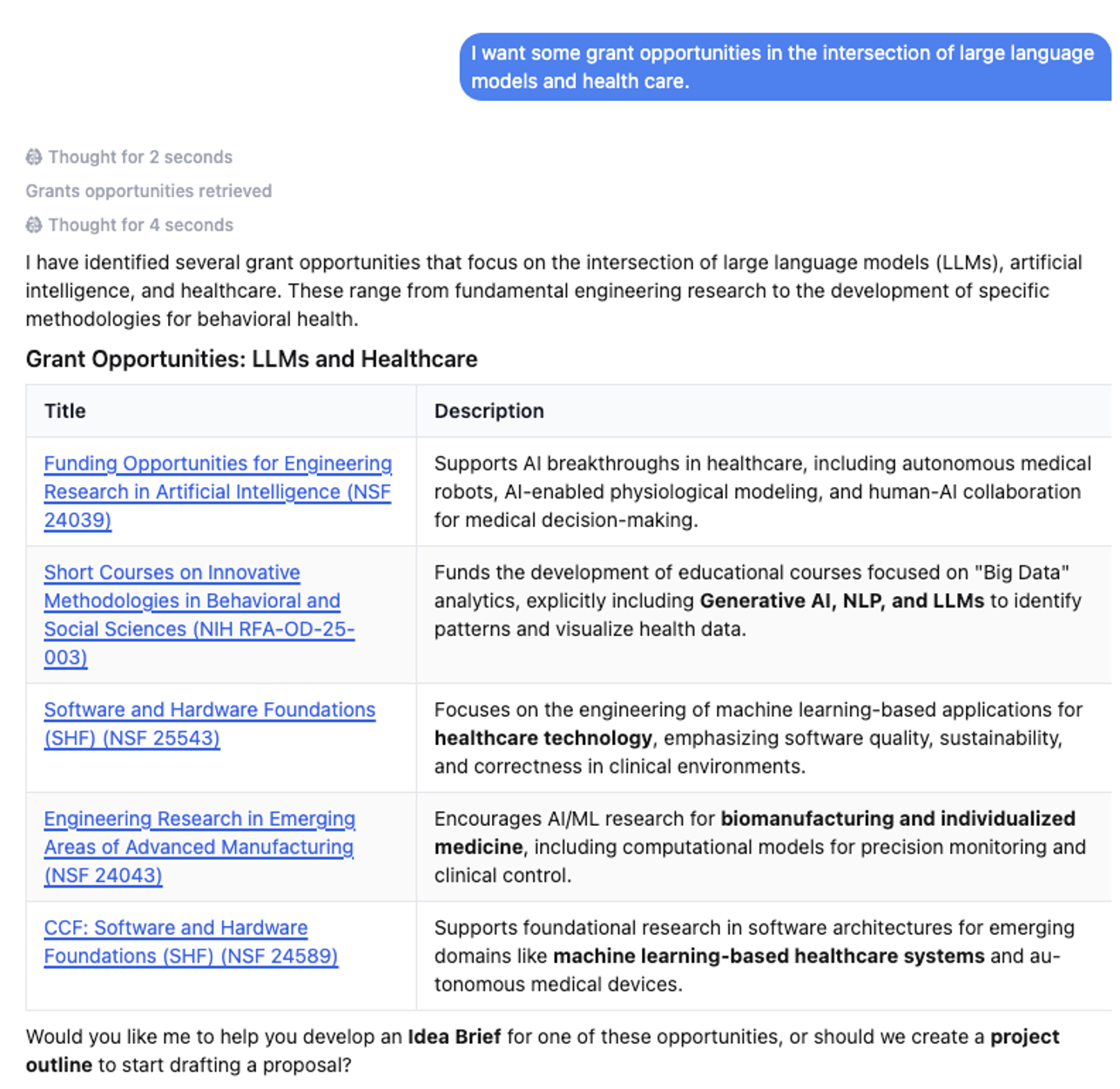}
\end{minipage}
\hfill
\begin{minipage}[t]{0.48\textwidth}
\centering
\textbf{Pane B: Google Search Results}\\[0.5em]
\includegraphics[width=0.95\textwidth]{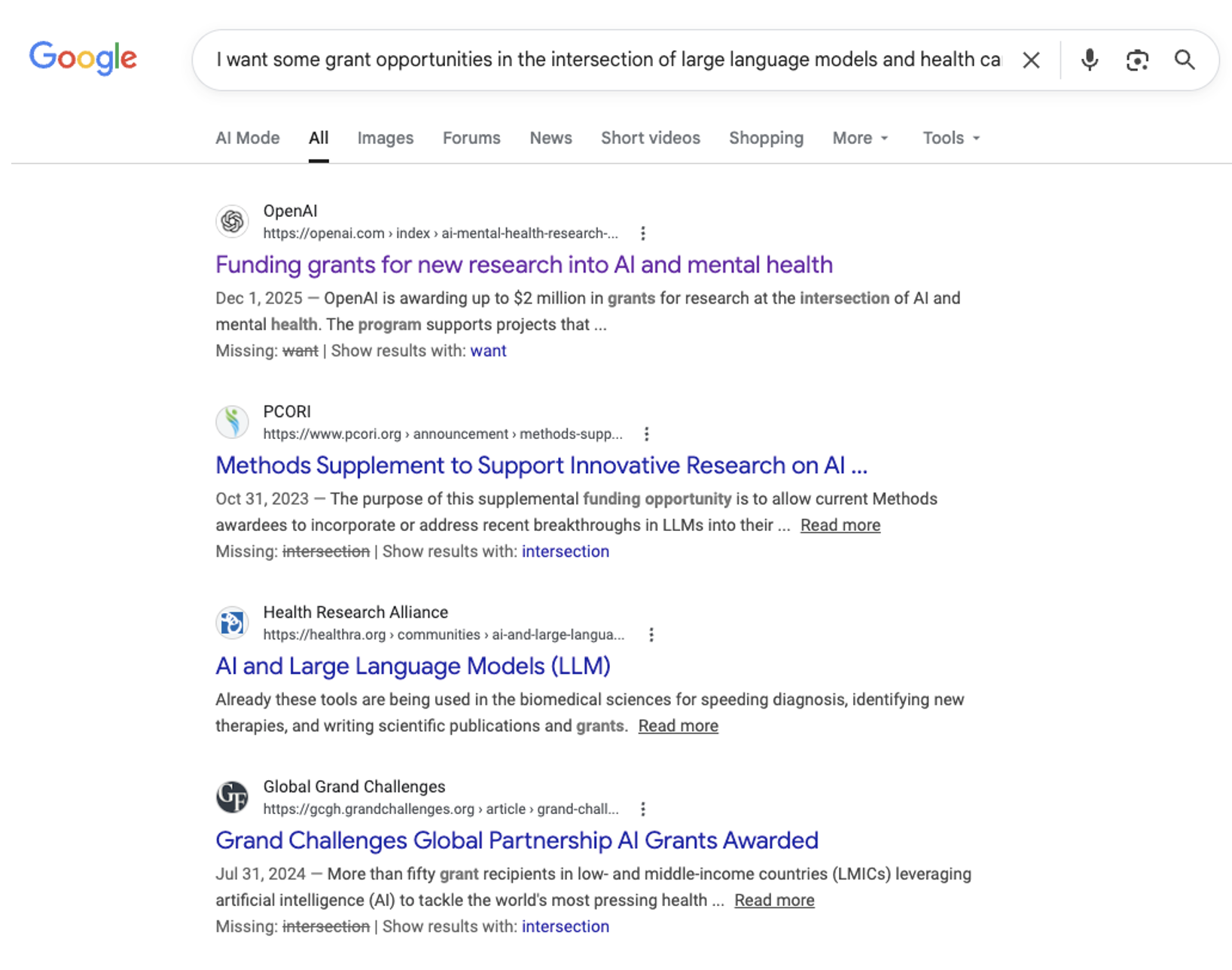}
\end{minipage}
\caption{Comparison of our conversational grant discovery system (Pane A) with traditional web search (Pane B). Our approach retrieves structured grant opportunities from the unified index (thereby avoiding hallucinations that could occur from direct LLM search), displaying relevant sources across multiple agencies with normalized metadata such as direct links and (truncated) deadlines. Traditional web search (Pane B) requires manual filtering through fragmented agency portals and lacks the opportunity data necessary for efficient grant discovery.}
\label{fig:comparison}
\end{figure*}

Consider a researcher in climate-adapted agriculture who uploads a grant proposal draft to the system. The agent automatically extracts domain-relevant keywords from the PDF and invokes \texttt{search\_index}, returning relevant opportunities (NSF, NIFA, USDA, DOE, NOAA, Gates Foundation) within 2 seconds, with results streaming incrementally. The researcher then iteratively refines: ``Which have deadlines more than six months away?'' and ``Do any support multi-institutional collaborations?'' A query about recent postings triggers \texttt{web\_search}, surfacing newly announced opportunities. This session—moving from broad discovery to a targeted shortlist of relevant programs—takes under 10 minutes. The equivalent manual search across NSF FastLane, NIFA, Grants.gov, and foundation portals typically consumes 30--45 minutes. Figure~\ref{fig:comparison} illustrates the comparison between our unified grant discovery system and traditional keyword search approaches.

\paragraph{Streaming and Transparency}

All agent reasoning is fully streamed to the client in real time. As the agent extracts keywords, queries the index, ranks results, and generates summaries, intermediate steps flow to the frontend immediately rather than being batched and returned at the end. This streaming architecture serves two critical purposes: (i) \textit{Responsiveness}: Users see progress immediately as the system works through each step, reducing perceived latency and supporting a natural conversational rhythm. (ii) \textit{Transparency}: Users can observe how the agent interpreted their research description and why specific opportunities were retrieved. This visibility enables appropriate calibration of trust—users develop understanding of both the agent's capabilities and its failure modes, supporting better decision-making about which results to pursue.

\section{Real-World Usage and Lessons Learned}

The unified index consolidates opportunities that researchers would otherwise discover through separate portals and databases. This consolidation provides several practical benefits: it reduces cognitive switching overhead that researchers incur when navigating between NSF FastLane, NIH REPORTER, Grants.gov, and individual foundation websites. Federated search systems address exactly this challenge by aggregating results from multiple heterogeneous sources into a unified query interface~\cite{shokouhi2011federated, wang2024feb4rag}. The single interface enables cross-domain discovery—a researcher in biomedical engineering might discover relevant opportunities in materials science or computational methods that align with their work, opportunities they might miss when searching databases in isolation.

The design of the conversational interface reflects observed patterns in how researchers explore funding. Rather than conducting a single, definitive search, researchers iteratively refine their queries to understand the landscape. They ask questions like ``Which of these programs support international collaborations?'' or ``Are there programs with later deadlines?'' or ``Show me foundation funding alongside federal programs.'' Conversational search systems have been shown to support this exploratory pattern by enabling multi-turn interactions that progressively constrain the search space~\cite{radlinski2017theoretical, vtyurina2017exploring}. The conversational interface naturally supports this iterative refinement, where each follow-up question applies an additional constraint or shifts focus without requiring the researcher to restart from their original query~\cite{avula2022effects}.

Key lessons from real-world deployment include: (1) heterogeneous data quality across sources requires graceful degradation rather than strict schema enforcement; (2) keyword extraction for funding discovery benefits significantly from iterative refinement when initial results are sparse or ambiguous; and (3) conversational interfaces must expose intermediate reasoning to enable users to understand and correct the agent's interpretations. The system handles edge cases robustly: when no results are found, the agent suggests alternative keyword refinements; when a grant is outdated, direct links redirect researchers to current opportunities on agency sites; when queries are out-of-scope, the agent transparently indicates limitations.

\section{Live Demonstration Scenario}\label{demo}

At the conference demonstration, attendees will interact with a live system showing grant discovery in action. We will present two primary scenarios:

\textbf{Scenario 1: PDF Upload with Automatic Keyword Extraction.} Attendees will upload a sample research paper or grant proposal (or provide one from our library). The system immediately displays the extracted keywords and domain tags, demonstrating that the agent understood the research without requiring manual specification. Within 2 seconds, the first results appear and stream incrementally, showing relevant grant opportunities from the unified index alongside brief summaries and direct links to agency pages. This scenario highlights the most compelling feature: researchers can ``drag and drop'' their research descriptions and immediately receive a curated list of opportunities.

\textbf{Scenario 2: Iterative Refinement and Web Search Complementarity.} Starting from the results of Scenario 1, attendees will observe the agent's response to follow-up queries: filtering by deadline, funding amount, or collaboration requirements. A critical moment occurs when an attendee asks, ``Are there any programs posted in the last week?'' The system will invoke \texttt{web\_search} to supplement the biweekly index, demonstrating how the hybrid approach captures relevant data across agencies (see Figure~\ref{fig:comparison}, Pane A) in contrast with standard web search where some of the results even in the top 5 are not completely relevant to the user (Pane B). Attendees will observe in real time the agent's decision logic: why certain tools were invoked, and the ranking of results.

Throughout both scenarios, attendees will see the streaming results, intermediate reasoning, and the quality of grant matching.

\paragraph{Availability}
The system is accessible at \texttt{https://grail.page} and requires no installation. Researchers can begin discovering grants immediately through their web browser. We welcome feedback from the community for improving the system.

\bibliographystyle{ACM-Reference-Format}
\bibliography{sample-base}

\end{document}